\documentclass[10pt,twocolumn,letterpaper]{article}
\usepackage{titling}

\usepackage{wacv}
\usepackage{times}
\usepackage{epsfig}
\usepackage{graphicx}
\usepackage{amsmath}
\usepackage{amssymb}
\usepackage{multirow}
\usepackage{pifont}
\usepackage{url}
\usepackage{sidecap}
\usepackage[numbers,sort&compress]{natbib}
\newcommand{\cmark}{\ding{51}}%
\newcommand{\xmark}{\ding{55}}%
\usepackage{hyperref}
\hypersetup{
    colorlinks=true,
    urlcolor=blue,
}
\wacvfinalcopy 

\def\wacvPaperID{354} 

\def\L{\mathcal{L}}
\def\T{\mathcal{T}}
\def\a{\mathbf{a}}
\def\v{\mathbf{v}}
\def\t{\mathbf{t}}
\def\x{\mathbf{x}}
\DeclareMathOperator*{\argmin}{arg\,min}
\def\ra{$\rightarrow$\ }

\def\vs{vs\onedot}
\def\eg{eg\onedot}
\def\ie{ie\onedot}
\def\cf{cf\onedot}

\begin{document}

\title{Coordinated Joint Multimodal Embeddings for Generalized Audio-Visual Zeroshot Classification and Retrieval of Videos}

\author{Kranti Kumar Parida\\
IIT Kanpur\\
\and
Neeraj Matiyali \\
IIT Kanpur\\
\and 
Tanaya Guha\\
University of Warwick\\
\and
Gaurav Sharma\\
NEC Labs America\\
}

\maketitle

\begin{abstract}
   We present an audio-visual multimodal approach for the task of zeroshot learning (ZSL) for classification and retrieval of videos. ZSL has been studied extensively in the recent past but has primarily been limited to visual modality and to images. We demonstrate that both audio and visual modalities are important for ZSL for videos.
Since a dataset to study the task is currently not available, we also construct an appropriate multimodal dataset with $33$ classes containing $156,416$ videos, from an existing large scale audio event dataset.
We empirically show that the performance improves by adding audio modality for both tasks of zeroshot classification and retrieval, when using multimodal extensions of embedding learning methods. We also propose a novel method to predict the `dominant' modality using a jointly learned modality attention network. 
We learn the attention in a semi-supervised setting and thus do not require any additional explicit labelling for the modalities. We provide qualitative validation of the modality specific attention, which also successfully generalizes to unseen test classes.
\end{abstract}

\section{Introduction}
Zeroshot learning (ZSL) refers to the setting when test time data comes from classes that were not seen during training. In the past few years, ZSL for classification has received significant attention \cite{akata2015evaluation, xian2018feature, chao2016empirical, frome2013devise, romera2015embarrassingly, lampert2014attribute, norouzi2013zero, akata2016label, verma2018generalized} due to the challenging nature of the problem, and its relevance to real world settings, where a trained model deployed in the field may encounter classes for which no examples were available during training. Initially, ZSL was proposed and studied in the setting where the test examples were from unseen classes and were classified into one of the unseen classes only \cite{lampert2014attribute}. This however is an artificial/controlled setting. More recent ZSL works thus focus on a setting where unseen test examples are classified into both seen and unseen classes \cite{xian2018zero, xian2018feature, verma2018generalized}.
The present work follows the latter setting known as the \emph{Generalized ZSL}. 

The majority of work involving generalized ZSL \cite{chao2016empirical, xian2018zero} has (i) worked with images, and (ii) used only visual representations along with text embeddings of the classes. When dealing with images, this is optimal. However, for the task of video ZSL, the audio modality, if available, may help with the task by providing complementary information. Ignoring the audio modality completely might even render an otherwise easy classification task difficult, \eg if we are looking to classify an example from the `dog' class, the dog might be highly occluded and not properly visible in the video, but the barking sound might be prominent.

\begin{figure}
    \centering
    \includegraphics[width=\columnwidth]{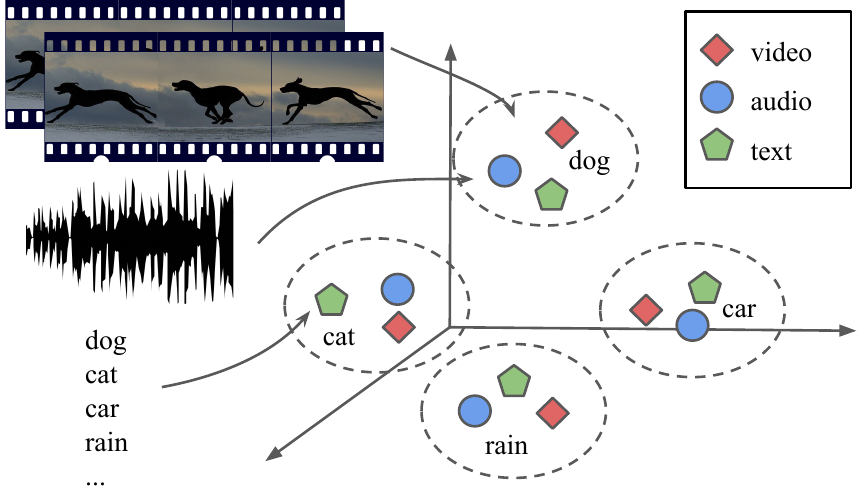}
    \caption{Illustration of the proposed method. We jointly embed all videos, audios and text labels into the same embedding space. We learn the space such that the corresponding embedding vectors for the same classes have lower distances than those of different classes. Once embeddings are learned, ZSL classification and crossmodal retrieval can be posed as a nearest neighbor search in the embedding space.}
    \label{fig:illus}
\end{figure}

In this work, we study the problem of ZSL for videos with general classes like, `dog', `sewing machine', `ambulance', `camera', `rain', and propose to use audio modality in addition to the visual modality. ZSL for videos is relatively less studied, \cf ZSL for images. There are several works on video ZSL for the specific task of human action recognition 
\cite{mishra2018generative, qin2017zero, xu2016multi}
but they ignored the audio modality as well. Our focus here is on leveraging both audio and video modalities to learn a joint projection space for audio, video and text (class labels). In such an embedding space, ZSL tasks can be formulated as nearest neighbor searches (fig.~\ref{fig:illus} illustrates the point). When doing classification, a new test video is embedded into the space and the nearest class embedding is predicted to be its class. Similarly, when doing retrieval, the nearest video or audio embeddings are predicted to be its semantic retrieval outputs. 

We propose cross-modal extensions of the embedding-based ZSL approach based on triplet loss for learning such a joint embedding space. We optimize an objective based on (i) two cross-modal triplet losses, one each for ensuring compatibility between the text (class labels) and the video, and the text and the audio, and (ii) another loss based on crossmodal compatibility of the audio and visual embeddings. While the triplet losses encourage the audio and video embeddings to come closer to respective class embeddings in the common space, the audio-visual crossmodal loss encourages the audio and video embeddings from the same sample to be similar. These losses together ensure that the three embeddings of the same class are closer to each other relative to their distance from those of different classes. 
The crossmodal loss term is an $\ell_2$ loss, and uses paired audio-video data, the annotation being trivially available from the videos. While the text-audio and text-video triplet losses use class annotations available for the seen classes during training, the crossmodal term uses the trivial constraint that audio and video from the same example are similar. 

As another contribution, we also propose a modality attention based extension, which first seeks to identify the `dominant' modality and then makes a decision based on that modality only if possible. To clarify our intuition of `dominant', we refer back to the dog video example  above, where the dog may be occluded but barking is prominent. In this case, we would like the audio modality to be predicted as dominant, and subsequently be used to make the class prediction. In case the attention network is not able to decide a clear dominant modality the inference then continues using both the modalities. This leads to a more interpretable model which can also indicate which modality it is basing its decision on. Furthermore, we show empirically that using such attention learning improves the performance, and brings it to be competitive to model trained on a concatenation of both modality features.

A suitable dataset was not available for the task of audio-visual ZSL. Hence, we construct a multimodal dataset with class level annotations. The dataset is a subset of a recently published large scale dataset, called Audioset~\cite{gemmeke2017audio}, which was primarily created for audio event detection and maintains a comprehensive sound vocabulary. We subsample the dataset to allow studying the task of audiovisual ZSL in a controlled setup. In particular, the subsampling ensures that (i) the classes have relatively high number of examples, with the minimum number of examples in any class being $292$, (ii) the classes belong to diverse groups, \eg animals, vehicles, weather events, (ii) the set of unseen classes is such that the pre-trained video networks could be used without violating the zeroshot condition, \ie the pre-training did not involve classes close to the unseen classes in our dataset. We provide more details in sec.~\ref{sec:dataset}.

In summary, our contributions are as follows. (i) We introduce the problem of audiovisual ZSL for videos, (ii) we construct a suitable dataset to study the task, (iii) we propose a multimodal embedding based ZSL method for classification and crossmodal retrieval, (iv) we propose a modality attention based method, which indicates which modality is dominant and was used to make the decision. We thoroughly evaluate our method on the dataset and show that considering audio modality, whenever appropriate, helps video ZSL tasks. We also show our method on standard ZSL datasets and results for some existing ZSL approaches for single-modality in our dataset as well. We also present qualitative results highlighting the improved cases using the proposed methods.

\section{Related Work}
\noindent\textbf{Zeroshot learning.} 

ZSL has been quite popular for image classification
\cite{socher2013zero, xian2016latent, akata2015evaluation, xian2018feature, chao2016empirical, frome2013devise, romera2015embarrassingly, lampert2014attribute, norouzi2013zero, akata2016label, verma2018generalized, kodirov2017semantic}, and recently has been used for object detection in images as well \cite{bansal2018zero, zhu2018zero, rahman2018zero}. The problem has been often addressed as a task of embedding learning, where the images and their class labels are embedded in a common space. The two types of class embeddings commonly used in the literature are based on (i) attributes like shape, color, and pose \cite{lampert2014attribute, romera2015embarrassingly, xian2018feature, verma2018generalized}, and (ii) semantic word embeddings \cite{frome2013devise, norouzi2013zero, xian2018feature, verma2018generalized}. Few works have also used both the embeddings together \cite{akata2016label, xian2016latent, akata2015evaluation}. Different from embedding learning, few recent works \cite{xian2018feature, verma2018generalized} have proposed to generate the data for the unseen class using a generative approach conditioned on the attribute vectors. The classifiers are then learned using the original data for the seen classes and the generated data for the unseen classes. This line of work follows the recent success of image generation methods \cite{goodfellow2014generative, reed2016generative}.
The initially-studied setting in ZSL refers to the one where the test examples were classified into unseen test classes only \cite{lampert2014attribute}. However, more recently the generalized version was proposed where they are classified into both seen and unseen classes \cite{chao2016empirical}. We address this later more practical, setting\footnote{Some earlier video retrieval works were called zeroshot, however, they are not strictly zeroshot in the current sense. Kindly see Supplementary material for a detailed discussion}.

Work on ZSL involving audio modality is scarce. We are aware of only one very recent work, where the idea of ZSL has been used to recognize unseen phonemes for multilingual speech recognition \cite{li2019zeroshot}.

\vspace{0.5mm} 
\noindent\textbf{Audiovisual learning.} 
In the last few years, there has been a significant growth in research efforts that leverage information from audio modality to aid visual learning tasks and vice-versa. Audio modality has been exploited for applications such as, audiovisual correspondence learning \cite{aytar2016soundnet, owens2016ambient, arandjelovic2017look, owens2018audio}, audiovisual source separation \cite{zhao2018sound, gao2018learning} and source localization \cite{arandjelovic2017objects, parekh2018identify, senocak2018learning}. Among the representative works, Owens et al.~\cite{owens2016ambient} used CNNs to predict, in a self-supervised way, if a given pair of audio and video clip is temporally aligned or not. The learned representations are subsequently used to perform sound source localization, and audio-visual action recognition. In a task of crossmodal biometric matching, Nagrani et al.~\cite{nagrani2018seeing} proposed to match a given voice sample against two or more faces. Arandjelovic et al.~\cite{arandjelovic2017objects} introduced the task of audio-visual correspondence learning, where a network comprising visual and audio subnetworks was trained to learn semantic correspondence between audio and visual data. Along the similar lines, Arandjelovic et al.~\cite{arandjelovic2017look} and Sencoak et al.~\cite{senocak2018learning} investigated the problem of localizing objects in an image corresponding to a sound input. Gao et al.~\cite{gao2018learning} proposed a multi-instance multilabel learning framework to address the audiovisual source separation problem, where they extract different audio components and associate them with the visual objects in a video. Ephrat et al.~\cite{ephrat2018looking} proposed a join audiovisual model to address the classical cocktail party problem (blind speech source separation).  Zhao et al.~\cite{zhao2018sound} proposed a self-supervised learning framework to address the problem of pixel-level (audio) source localization \cite{kidron2005pixels}.

\section{Coordinated Joint Multimodal Embeddings}
\label{sec:approach}

\begin{figure}
    \centering
    \includegraphics[width=\columnwidth]{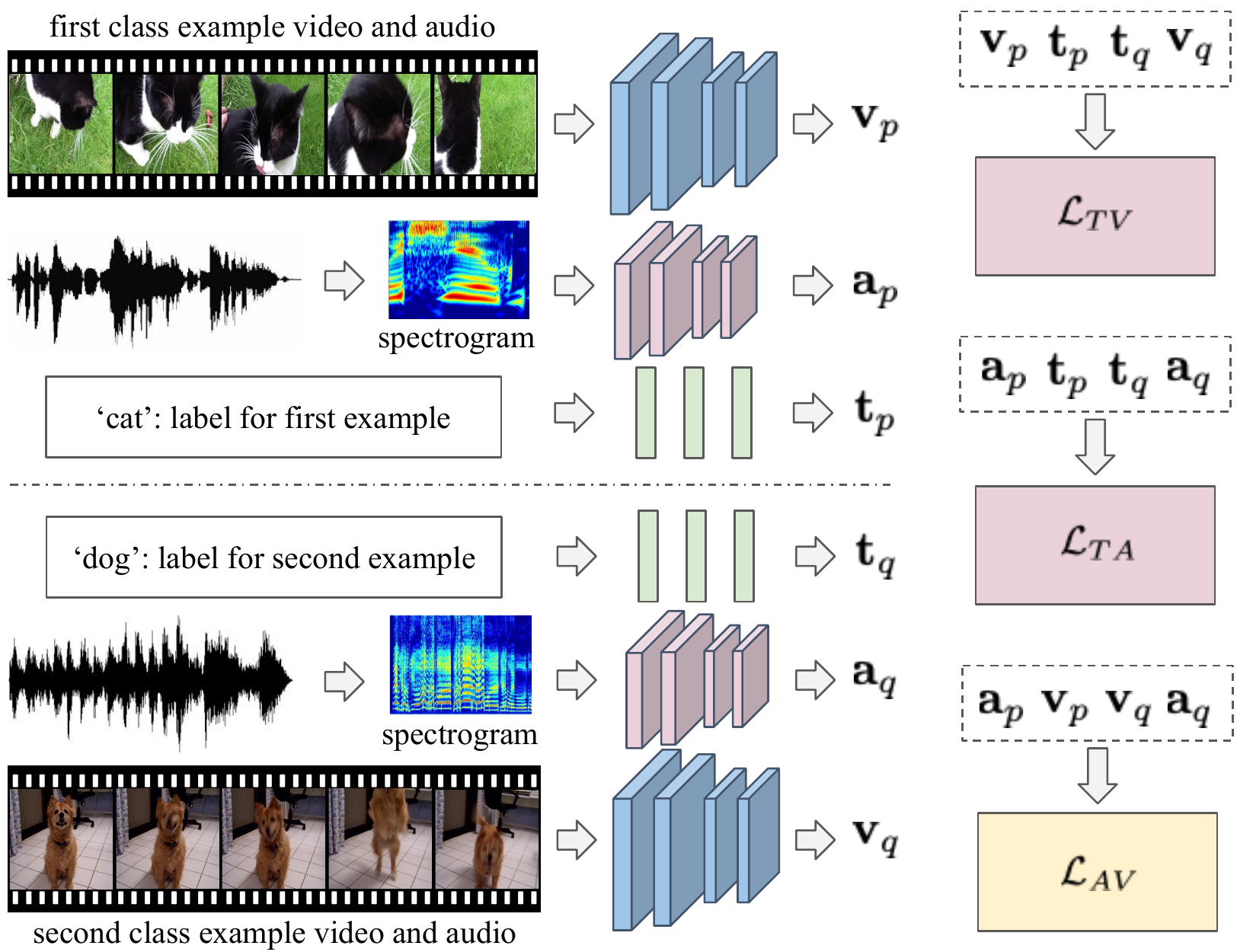}
    \caption{Block diagram of the proposed approach. Pairs of video, audio and text networks share weights.}
    \label{fig:approach}
\end{figure}

We now present our method in detail. Fig.~\ref{fig:illus} illustrates the basic idea and Fig.~\ref{fig:approach} gives the high level block diagram of the proposed method.
Our method works by projecting all three inputs, audio, video and text, onto a common embedding space such that class constraints and crossmodal similarity constraints are satisfied. The class constraints are enforced using bimodal triplet losses between audio and text, and video and text embeddings. Denoting $\a_i, \v_i, \t_i$ as the audio, video and text embedding (we explain how we obtain them shortly) for an example $i$, we define the bimodal triplet losses as follows
\begin{align}
    \small
    \L_{TA}(\a_p, \t_p, \a_q, \t_q) &= \left[ d(\a_p, \t_p) - d(\a_q, \t_p) + \delta \right]_+\\
    \L_{TV}(\v_p, \t_p, \v_q, \t_q) &= \left[ d(\v_p, \t_p) - d(\v_q, \t_p) + \delta \right]_+
\end{align}
where, $(\a_p, \v_p, \t_p)$ and $(\a_q, \v_q, \t_q)$ are two example videos with $\t_p\neq \t_q$. These losses force the audio and video embeddings to be closer to the correct class embedding by a margin $\delta>0$ \cf the incorrect class embeddings.

We also use a third loss to ensure the crossmodal similarity between the audio-video streams that come from the same video in the common embedding space. This loss is simply a $\ell_2$ loss given by
\begin{align}
    \L_{AV}(\a_p, \v_p) = \| \a_p - \v_p \|_2^2.
\end{align}
The full loss function is thus a weighted average of these three losses.
\begin{align}
    \L & = \lambda \sum_{p \in \T} \L_{AV} + 
    \gamma \sum_{\substack{p,q \in \T\\ y_p \neq y_q}} 
    \left\{ 
        \alpha_v \L_{TV}  + \alpha_a \L_{TA}  \right\},
    \label{eqn:finalobj}
\end{align}
where, $\lambda, \gamma, \alpha_v, \alpha_a$ are the hyperparameters that control the contributions of the different terms, and $\T$ is the index set over the training examples $\{(\a_i, \v_i, y_i)| i=1,\ldots,N\}$ with $y_i$ being the class label.
With these three losses over all pairwise combinations of the modalities, \ie $\L_{TV}, \L_{TA}, \L_{AV}$, we force the embeddings from all the three modalities to respect the class memberships and similarities.
\vspace{0.5em} \\
\textbf{Representations and parameters.} We now need to specify the parameters over which these losses are optimized. We represent each of the three types of inputs, \ie audio, video, and text, using the corresponding state-of-the-art neural networks outputs which we denote as $f_a(\cdot), f_v(\cdot), f_t(\cdot)$. We project each representation with corresponding neural networks which are small MLPs, denoted as $g_a(\cdot), g_v(\cdot), g_t(\cdot)$ with parameters $\theta_a, \theta_v, \theta_t$ (we give details about all these networks in the implementation details sec.~\ref{sec:experiments}). Finally, the representations are obtained by passing the input audio/video/text through the corresponding networks sequentially, \ie $\x = g_x \circ f_x (X)$ where $x \in {\a,\v,\t}$ and $X$ is the corresponding raw audio/video/text input. We keep the initial network parameters fixed to be that of the pretrained networks and optimize over the parameters of the projection networks. Hence, the full optimization is given as,
\begin{align}
    \theta_a^*, \theta_v^*, \theta_t^*
    & = \argmin_{\theta_a, \theta_v, \theta_t} \L(\T).
\end{align}
We train for the parameters using standard backpropagation for neural networks.
\vspace{0.5em} \\
\textbf{Inference.}
Once the model has been learned, we use nearest neighbor in the embedding space for making predictions. In the case of classification, the audio and video are embedded in the space and the class embedding with the minimum average distance with them is taken as the prediction, \ie 
\begin{align}
    \t^* = \argmin_\t \left\{d(\a, \t) + d(\v, \t)\right\}.
\end{align}
In the case of (crossmodal) retrieval, the sorted list of audio or video examples are returned as the result, based on their distance from the query in the embedding space.
\vspace{0.5em} \\
\textbf{Modality attention based learning.}
In the prequel, the method learns to make a prediction (classification or retrieval) using both the audio as well as video modalities. We augment our method to predict modality attention to find the dominant modality for each sample, \eg in case when the object is occluded or not visible, but the characteristic sound is clearly present we want the network to be able to make the decision based on the audio modality only. We incorporate such attention by adding a attention predictor network $f_{attn}(\cdot)$, with parameters $\theta_{attn}$, which takes the concatenated audio and video features as inputs and predicts a scalar $\alpha$ which gives us the relative importance weights $\alpha_v = \alpha, \alpha_a = 1 - \alpha$ in eq.~\ref{eqn:finalobj}. All the network parameters are then learned jointly.

To further guide the attention network, we use the intuition that when one modality is dominant, say audio, the correct class embedding is expected to be much closer to the audio embedding, than the other classes \cf the video embedding. Hence the entropy of the prediction probability distribution over classes, for the dominant modality, should be very low. To compute such distribution, we first compute the inverse of the distances of the query embedding to all the class embeddings, and then $\ell_1$ normalize the vector. We then derive a supervisory signal for $\alpha$ using the entropies computed \wrt audio and video modalities, denoted $e_a, e_v \in [0, \log N_c]$ where $N_c$ is the number of classes over which prediction is being done, as
\begin{align}
    \alpha = \left\{ 
                \begin{array}{ll}
                    0, & \textrm{if } e_v < e_a - \xi \\
                    1, & \textrm{if } e_a < e_v - \xi \\
                    0.5, & \textrm{otherwise}
                \end{array}
            \right.
\label{eqn:alpha}
\end{align}
where, $\xi>0$ is a threshold parameter for preferring one of the modalities based on their entropy difference. The modality attention objective becomes, $\L_{attn} = \L + \L_{CE-\alpha}$, where $\L$ is the objective from eq.~\ref{eqn:finalobj}, $\L_{CE-\alpha}$ is the cross entropy loss on $\alpha$ based on the generated supervision above. This loss is then minimized jointly over all $\theta_a, \theta_v, \theta_t, \theta_{attn}$.
\vspace{0.5em} \\
\textbf{Modality selective inference with attention.} While attention is interesting at training as it helps identify the dominant modality and learn better models. We also use attention to make inference using only the predicted dominant modality at test time. When the predicted attention is higher than a threshold for one of the modalities we only compute distance for that modality in the embedding space and use that to make the prediction. 

We could also use the above computed $\alpha$ value based on the difference of entropies of the prediction distributions (eq.~\ref{eqn:alpha}) at test time, even when not training with modality attention. We use that as a baseline to verify that learning to predict the attention helps improve the performance.
\vspace{0.5em} \\
\textbf{Calibrated stacking in generalized ZSL (GZSL).} The common problem with GZSL setting is that the classifier is always biased to wards the seen classes. This reduces the performance for the unseen classes as the unseen examples are often misclassified to one of the seen classes. A simple approach to handle this was proposed in \cite{chao2016empirical}, where the authors suggested to reduce the scores for the seen classes. The amount $\beta$ by which the scores are additively reduced for the seen classes, is a parameter which needs to be tuned. We use the approach of calibrated stacking, and as we are working with distances instead of similarities, we use the modified prediction rule at inference, given by
\begin{align}
    \t^* = \argmin_{\t, c \in \{S+U\}} \left\{d_c(\x, \t) + \beta \mathcal{I}(c \in S)\right\},
\end{align}
where, $\x$ can be audio, video or concatenated feature, $\mathcal{I}$ is the indicator function which is $1$ when the input condition is true and $0$ otherwise. 
\section{Proposed AudioSet ZSL Dataset}
\label{sec:dataset}

A large scale audio dataset, AudioSet \cite{gemmeke2017audio}, was recently released containing segments from in-the-wild YouTube videos (with audio). These videos are weakly annotated with different types of audio events ranging from human and animal sounds to musical instruments and environmental sound. In total, there are $527$ audio events, and each video segment is annotated with multiple labels.

The original dataset being highly diverse and rich, is often used in parts to address specific tasks \cite{gao2018learning,2017dcase}. 

\begin{figure}
    \centering
    \includegraphics[width=.95\columnwidth]{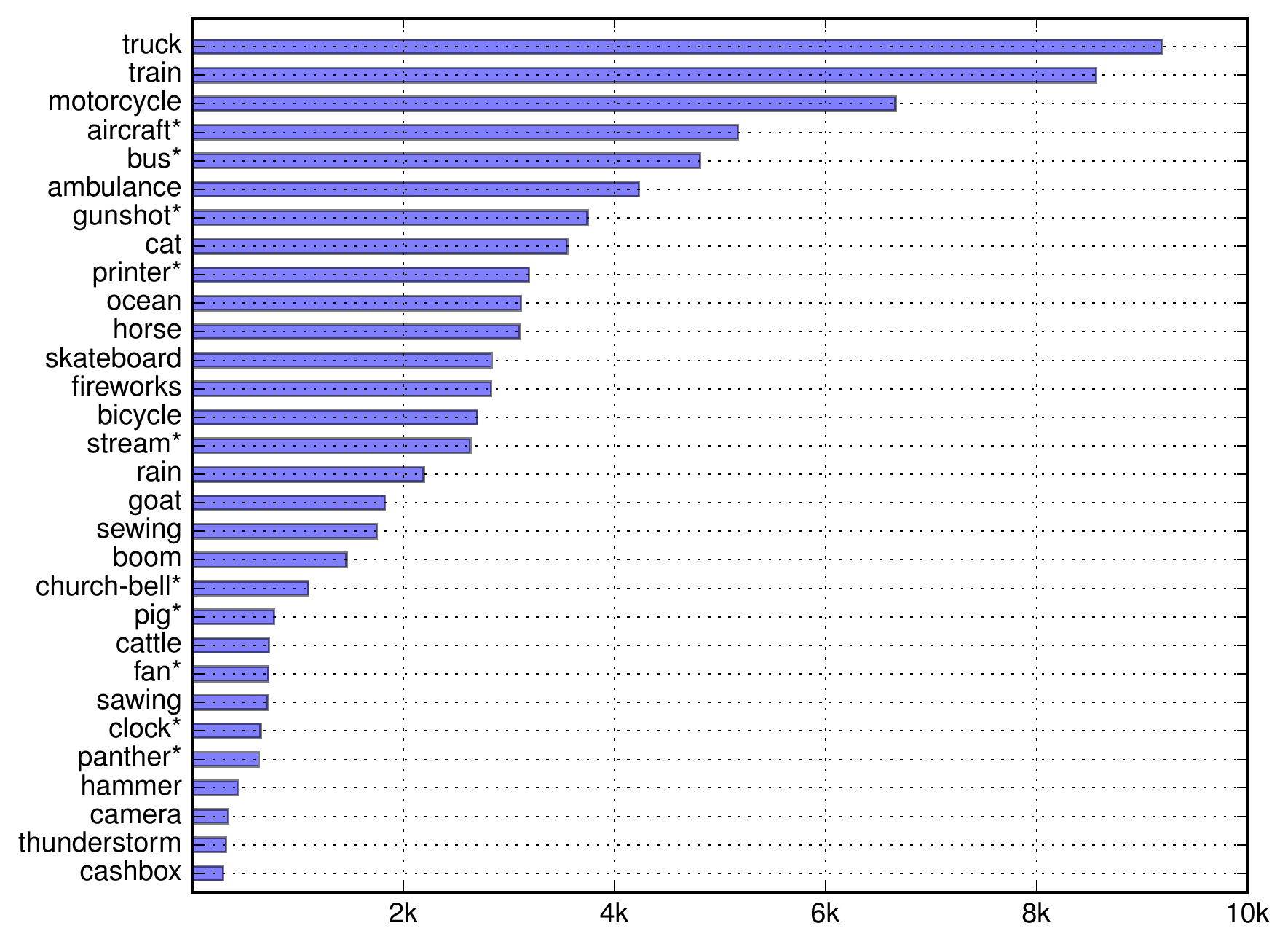}    \caption{Distribution of the different classes in \texttt{AudioSetZSL}. Apart from these three other classes included in the dataset are dog, bird and car containing $12646$, $25153$ and $38315$ examples. The unseen classes are appended with a `*'.}
    \label{fig:dataset}
\end{figure}

To study the task of audiovisual ZSL, we construct a subset of the Audioset containing $156416$ video segments.  We refer to this subset as the \texttt{AudioSetZSL}. While the original dataset was multilabel, the example videos were selected such that every video in \texttt{AudioSetZSL} has only one label, \ie it is a multiclass dataset. Fig.~\ref{fig:dataset} shows the number of examples for different classes in \texttt{AudioSetZSL}, tab.~\ref{table:dataset_stats} gives some statistics. 

We follow the steps below to create the \texttt{AudioSetZSL}: (i) We remove classes with confidence score (for annotation quality) less than $0.7$, (ii) we then determine the group of classes that are semantically similar, e.g.~animals, vehicles, water bodies. We do so to ensure that the seen and unseen classes for ZSL have some similarities and the task is feasible with the dataset.
(iii) After selecting the classes, we discard 
highly correlated classes within those groups to have a challenging dataset, obtaining $33$ classes.
(iv) We then remove the examples which correspond to more than one of the $33$ classes to keep the dataset multiclass. We  then remove the examples that are no longer available on YouTube. Tab.~\ref{table:dataset_stats} and Fig.~\ref{fig:dataset} give some statistics and more details are in the supplementary document.

To create the \emph{seen, unseen} splits for ZSL tasks, we selected a total of $10$ classes spanning all the groups as the zero-shot classes (marked with `*' in Fig.~\ref{fig:dataset}). We ensure that the unseen classes have minimal overlap with the Kinetics dataset \cite{CarreiraCVPR2017} training classes as we use CNNs pre-trained on that. We do so by not choosing any class whose class embedding similarity is greater than $0.8$ with any of the Kinetics train class embeddings in the word2vec space.

\begin{table}
\centering
\begin{tabular}{c|c|c|c}
\hline
min & max & mean & std. dev. \\
\hline  \hline
292 & 38315 & 4739.88 & 7693.10\\
\hline
\end{tabular}
\caption{Statistics on the number of examples per class for the \texttt{AudioSetZSL} dataset.}
\label{table:dataset_stats}
\vspace{-1em}
\end{table}

We finally split, both the seen and unseen classes, as $60-20-20$ into train, validation and test sets. We set the protocol to be as follows. Train on the train classes and then test on seen class examples and unseen class examples, both being classified into one of all the classes. The performance measure is mean class accuracies for seen classes and unseen classes and the harmonic mean of these two values, following that in image based ZSL work \cite{xian2018zero}.

\section{Experiments}
\label{sec:experiments}
\noindent
\textbf{Implementation details}
The audio network $f_a(\cdot)$ is based on that of \cite{kumar2018knowledge}, and is trained on the spectrogram of the audio clips in the train set of our dataset. We obtain the audio features after seven \texttt{conv} layers of the network, and average them to obtain $1024$D vector. The video network, denoted as $f_v(\cdot)$ is an inflated 3D CNN network which is pretrained on the Kinetics dataset \cite{CarreiraCVPR2017} and a large video dataset of action recognition. We also obtain the  video features form the layer before the classification layer and average them to get a vector of $1024$D. Finally the text network, denoted as $f_t(\cdot)$ is the well known \texttt{word2vec} network pretrained on Wikipedia \cite{mikolov2018advances} with output dimension of $300$D.

The projection model for text embeddings was fixed to be a single layer network, where as for the audio and video was fixed to be a two layer network, with the output dimensions matching for all. In order to find the seen/unseen class bias parameter $\beta$ we divide the maximum and minimum possible value of $\beta$ into $25$ equal intervals and then evaluate performances on the \textit{val} set. We chose the best performing $\beta$ among those. 
\vspace{0.5em}\\
\textbf{Evaluation and performance metrics.}
We report the mean class accuracy (\% mAcc) for the classification task and the mean average precision (\% mAP) for the retrieval tasks. The performance for the seen classes (denoted as S) is classified (retrieved) over all the classes (seen and unseen), and that for the unseen classes (denoted as U) are also reported. The harmonic mean HM of S and U indicates how well the system performs on both seen and unseen categories on average. For classification, we classify each test example, and for retrieval, we perform leave-one-out testing, \ie each test example is considered as a query with the rest being the gallery. The performance reported is (mean class) averaged. 
\vspace{0.5em}\\
\textbf{Methods reported.}
We report performances of audio and video only methods, \ie only the respective modality is used to test and train. We also report a naive combination by concatenation of features from audio and video modalities before learning the projection to the common space. This method allows zeroshot classification and retrieval only when both the modalities are available, and it does not allow crossmodal retrieval at all. We then report performances with the proposed Coordinated Joint Multimodal Embeddings (CJME) method, when modality attention is used and when it is not used. In either of the cases, we can choose dominant modality (or not) based on the $\alpha$ value (eq.~\ref{eqn:alpha}). We report with both the cases.

We also compare our approach to two other baseline methods, namely pre-trained features and GCCA. In `Pre-trained' method the raw features obtained from the individual modality pre-trained network are directly used for retrieval as both are of same dimensions. This can be considered as one of the lower bound since no common projection is learned for the different modalities in this case. GCCA \cite{kettenring1971canonical} or Generalized Canonical Correlation Analysis is the standard extension of the Canonical Correlation Analysis (CCA) method from two-set method to multi-set method, where the correlation between the example pair from each sets are maximized. We use here the GCCA to maximize the correlation between all the three modalities (text, audio and video) for every example triplet in the dataset.

\subsection{Quantitative Evaluation}
\label{sec:quant}
\begin{figure}
\centering
	\includegraphics[width=0.49\columnwidth]{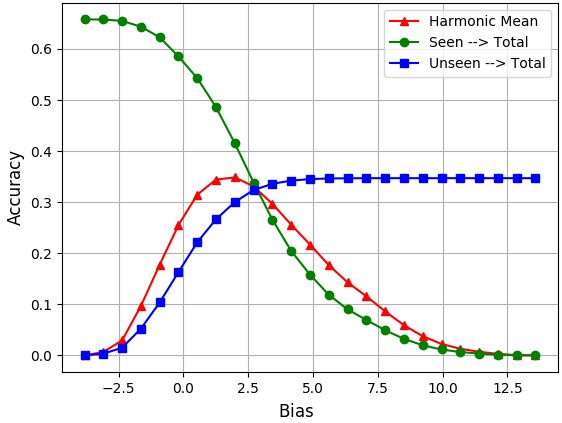}
	\includegraphics[width=0.49\columnwidth]{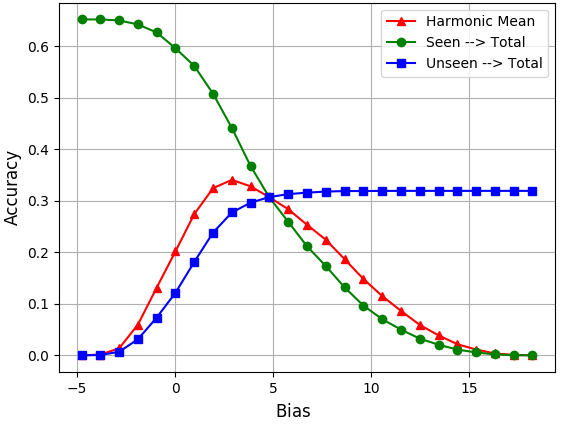}
\caption{Effect of classification performance for model M1 (left) and model M2 (right) with different values of bias parameter.
}
\label{fig:bias_corr}
\end{figure}

\noindent
\textbf{Evaluation of calibrated stacking performance.} 
We have shown the improvement in performance with the approach of calibrated stacking in
Fig.~\ref{fig:bias_corr}. This shows the performances with different values of the bias parameter, \ie accuracies for seen and unseen classes, as well as their harmonic means.
We observe that the performance increases with the initial increase in bias, and then falls after a certain point as expected. We choose the best performing value of the class bias on the \emph{val} set and then fix it for the experiments on \emph{test} set.

\begin{table}
\centering
\resizebox{.9\columnwidth}{!}{
\begin{tabular}{c|c|ccc}
\hline
Train Modality & Test Modality & S & U & HM \\
\hline \hline 
audio          & audio & 28.35 & 18.35 & 22.22 \\
CJME          & audio & 25.58  & 20.30 & 22.64 \\
\hline
video         & video & 43.27 & 27.11 & 33.34 \\
CJME  & video & 41.53 & 28.76 & 33.99 \\
\hline
both (concat) & both & 45.83 & 27.91 & 34.70 \\
CJME & both   & 30.29 & 31.30 & 30.79 \\
\hline
CJME (no attn) & audio or video & 31.72 & 26.31 & 28.76  \\
CJME (w/ attn) & audio or video & 41.07 & 29.58 & 34.39\\
\hline
\end{tabular}
}
\caption{Zeroshot classification performances (\% mAcc) achieved with audio only, video only, and both audio and video used for training and test. Note that the audio and video concatenation model requires both the modalities to be available during testing.}
\label{tab:zsc}
\vspace{-1 em}
\end{table}
\vspace{0.5em}
\textbf{Zeroshot audio-visual classification.}
Tab.~\ref{tab:zsc} gives the performances of the different models for the task of zeroshot classification. We make multiple observations here. The video modality performs better than the audio modality for the task ($33.34$ \vs $22.22$ HM), which is interesting as the original dataset was constructed for audio event detection. We also observe that when both audio and video modalities are used by simply concatenating the feature from the respective pre-trained networks, the performance increases to $34.70$. This shows that adding the audio modality is helpful for zeroshot classification. Our coordinated joint multimodal embeddings (denoted CJME in the table) improves the performance of video and audio only models on the respective test sets by modest but consistent margins. This highlights the efficacy of the proposed method to learn joint embeddings which are comparable (slightly better) than individually trained models.

The performance of the proposed method is lower without attention learning and selective modality based test time prediction \cf the concatenated input model ($30.79$ \vs $34.70$), but is comparable to it when trained and tested with attention ($34.39$). Also, when we do not train for attention but use selective modality based prediction the performance falls ($28.76$). Both these comparisons validate that the modality attention learning is an important addition to the base multimodal embedding learning framework.
\vspace{0.5em}\\
\textbf{Zeroshot audio-visual retrieval.}
Table~\ref{tab:zsr} compares the performances of different models for the task of zeroshot retrieval. 
The performance on the unseen classes are quite poor, albeit it is approximately three times the baseline pre-trained performance. This is because of the bias towards the seen classes in generalized ZSL. This happens for classification setting as well but is corrected for explicitly by reducing the scores of the seen classes. However, in a retrieval scenario, since the class of the gallery set member is not known in general, such correction can not be applied. We tried classifying the gallery sets first and then applying the seen/unseen class bias correction, however that did not improve results possibly because of erroneous classifications.

\begin{table}
\centering
\resizebox{.85\columnwidth}{!}{
\begin{tabular}{c|c|ccc}
\hline
Model & Test & S & U & HM \\
\hline \hline
pre-trained        & T\ra A & 3.83 &  1.66 & 2.32 \\
GCCA \cite{kettenring1971canonical}    & T\ra A & 49.84 &  2.39 & 4.56 \\
audio        & T\ra A & 43.16 & 3.34 & 6.20 \\
CJME & T\ra A & 48.24  &  3.32  &  6.21 \\
\hline
pre-trained         & T\ra V & 3.83 & 2.53 & 3.05 \\
GCCA \cite{kettenring1971canonical}    & T\ra V & 57.67 &  3.54 & 6.67 \\
video        & T\ra V & 48.62 & 5.25 & 9.47 \\
CJME & T\ra V & 59.39  & 5.55  & 10.15 \\
\hline
both (concat)& T\ra AV & 63.13 & 7.80 & 13.88 \\
CJME & T\ra AV & 65.45 & 5.40 & 9.97\\
\hline
CJME (no attn) & T\ra A or V & 65.74  &  5.09  &  9.45 \\
CJME (w/ attn) & T\ra A or V & 62.97 & 5.67 & 10.41\\
\hline
\end{tabular}
}
\caption{Zeroshot retrieval performances (\% mAP) achieved by models when audio only, video only, and both audio and video modalities are used for training and test. Note that the audio and video concatenation based model requires both modalities at test time also and can not predict using any single one.}
\label{tab:zsr}
\vspace{-1em}
\end{table}
We observe, from Tab.~\ref{tab:zsr}, similar trends as with zeroshot classification. The proposed CJME performs similar to audio only ($6.20$ \vs $6.21$) and slightly better than video only ($9.47$ \vs $10.15$) models but consistently outperforms both pre-trained and GCCA model. Compared to the audio and video features concatenated model, the performance without modality attention based training are lower ($13.88$ \vs $9.47$) which improve upon using attention at training ($10.41$), albeit staying a little lower \cf similar in the classification case. We thus conclude that CJME is as good as audio only or video only model and is competitive \cf concatenated features model, while allowing crossmodal retrieval, which we evaluate next.
\begin{table}
\centering
\resizebox{.9\columnwidth}{!}{
\begin{tabular}{c|c|ccc}
\hline
Model & Test & S & U & HM \\
\hline \hline
pre-trained & audio \ra video & 3.61 & 2.37 & 2.86  \\
GCCA \cite{kettenring1971canonical} & audio \ra video & 22.12 & 3.65  & 6.26 \\
CJME & audio \ra video & 26.87  &  4.31  &  7.43 \\
\hline 
pre-trained & video \ra audio & 4.22  & 2.57  & 3.19  \\
GCCA \cite{kettenring1971canonical} & video \ra audio & 26.68 & 2.98 & 5.26 \\
CJME & video \ra audio & 29.33  &  4.35  &  7.58 \\
\hline
\end{tabular}
}
\caption{Zeroshot crossmodal retrieval performances (\% mAP).}
\label{tab:zsr_cross}
\vspace{-1em}
\end{table}
\vspace{0.5em} \\
\textbf{Crossmodal retrieval.}
Since CJME learns to embed both audio and video modality in a common space, it allows for doing crossmodal retrieval from audio to video and vice-versa. Tab.~\ref{tab:zsr_cross} gives the performances of such crossmodal retrieval from audio and video domains. We observe that the retrieval accuracy in the case of crossmodal retrieval are $7.43$ and $7.58$ for audio to video and video to audio respectively. Due to the inability to do seen/unseen class bias correction, we observe a large gap between the retrieval performance of seen classes \cf unseen classes, which stays true in the case of crossmodal retrival as well. The performance is still three times better than the raw pre-trained features. We believe these are encouraging initial results on the challenging task of audio-visual crossmodal retrieval on real world unconstrained videos in zeroshot setting.
 \begin{table}[t]
 \centering
\resizebox{.75\columnwidth}{!}{
 \begin{tabular}{c|c|c||ccc}
 \hline
 $\L_{AV}$ & $\L_{TA}$& $\L_{TV}$ & S & U & HM \\
 \hline \hline
 \xmark & \xmark & \cmark  &  1.26  &  10.13  &  2.24  \\
 \xmark&   \cmark    &   \xmark   &  3.00  &  4.18  &  3.49  \\
 \xmark&   \cmark&   \cmark   &  31.20  &  28.47  &  29.77   \\
 \cmark &   \xmark   &   \cmark  &  30.39  &  27.31  &  28.76  \\
 \cmark &   \cmark &   \xmark  &  30.07  &  25.06  &  27.33  \\
 \cmark &   \cmark &   \cmark  &  33.29  &  28.18  &  30.53  \\
 \hline
 \end{tabular}
 }
 \caption{Ablation study to verify the contribution of different loss terms. Performances for proposed CJME (with attention) method on zeroshot classification (\% mAcc)}
 \label{tab:ablation}
 \vspace{-1em}
 \end{table}
 \vspace{0.5em} \\
\textbf{Ablation of the different loss components.} Tab.~\ref{tab:ablation} gives the performances in the different cases when we selectively turn off different combination of losses in the optimization objective eq.~\ref{eqn:finalobj}. We observe that all three losses contribute positively towards the performance. When either of the triplet loss is turned off, the performance drastically fall to $\sim3$, but when the crossmodal audio-video loss is added with one of the triplet losses turned off, they recover to reasonable values $\sim28$. Compared to the final performance of $30.53$, when the text-audio, text-video and audio-video losses are turned off, the performances fall to $28.76$, $27.33$ and $29.77$ respectively. Thus we conclude that each component in the loss function is useful and that the networks (which are already pre-trained on auxiliary classification tasks) need to be trained for the current task to give meaningful results.

\begin{table*}
\centering
\begin{minipage}[b]{0.66\textwidth}
\resizebox{\textwidth}{!} {
\renewcommand{\arraystretch}{1.25}
\begin{tabular}{c|ccc|ccc|ccc|ccc}
\hline
& \multicolumn{3}{c|}{\textbf{SUN}} & \multicolumn{3}{c|}{\textbf{CUB}} & \multicolumn{3}{c|}{\textbf{AWA1}}& \multicolumn{3}{c}{\textbf{AWA2}} \\
\textbf{Method} & \textbf{U} & \textbf{S} & \textbf{HM} & \textbf{U} & \textbf{S} & \textbf{HM} & \textbf{U} & \textbf{S} & \textbf{HM} & \textbf{U} & \textbf{S} & \textbf{HM}  \\ \hline \hline
CONSE~\cite{norouzi2013zero} & 6.8 & 39.9 & 11.6 & 1.6 & 72.2 & 3.1 & 0.4 &88.6 &0.8 & 0.5 & 90.6 & 1.0 \\
DEVISE~\cite{frome2013devise} & 16.9 & 27.4 & 20.9 & 23.8 & 53.0 & 32.8 & 13.4 &68.7 &22.4 & 17.1 & 74.7 & 27.8  \\
SAE~\cite{kodirov2017semantic} & 8.8 & 18.0 & 11.8 & 7.8 & 54.0 & 13.6 & 1.8 &77.1 &3.5 & 1.1 & 82.2 & 2.2 \\
ESZSL~\cite{romera2015embarrassingly} & 11.0 & 27.9 & 15.8 & 12.6 & 63.8 & 21.0 &6.6&75.6&12.1& 5.9 & 77.8 & 11.0 \\
ALE~\cite{akata2016label} & 21.8 & 33.1 & 26.3 & 23.7 & 62.8 & {34.4} & 16.8 & 76.1 & 27.5 & 14.0 & 81.8 & 23.9 \\ 
\hline
CJME & 30.2 &23.7 & {26.6} & 35.6 & 26.1 & 30.1 &   29.8& 47.9 & {36.7} &51.9 &36.8 & {43.1} \\
\hline
\end{tabular}
}
\caption{Comparison with existing methods on standard datasets (projection based methods only, see sec.~\ref{sec:sota} for details)}
\label{tab:zsl_otherdata}
\end{minipage}
\hfill
\begin{minipage}[b]{.32\textwidth}
\resizebox{\textwidth}{!} {
\begin{tabular}{c|cc|ccc}
\hline
&\multicolumn{2}{c|}{\textbf{Modality}}&&&\\
\textbf{Method}& train & test   &\textbf{S} & \textbf{U} & \textbf{HM} \\
\hline \hline 
CONSE~\cite{norouzi2013zero}  & video & video &48.5 & 19.6 & 27.9  \\
DEVISE~\cite{frome2013devise} & video & video &39.8 & 26.0 & 31.5  \\
SAE~\cite{kodirov2017semantic}  & video & video &29.3 & 19.3 & 23.2 \\
ESZSL~\cite{romera2015embarrassingly} & video & video & 33.8 & 19.0 & 24.3 \\
ALE~\cite{akata2016label} & video & video & 47.9 & 25.2 & 33.0 \\
\hline
CJME & video & video & 43.2 & 27.1 & 33.3\\
CJME & both & video & 41.5 & 28.8 & 33.9 \\
CJME & both & both & 41.0 & 29.5 & {34.4} \\
\hline
\end{tabular}
}
\caption{Comparison with existing methods on proposed dataset (projection based methods only, see sec.~\ref{sec:sota} for details)}
\label{tab:zsl_owndata}
\end{minipage}
\end{table*}

\begin{figure*}
\centering

\includegraphics[width=0.97\textwidth]{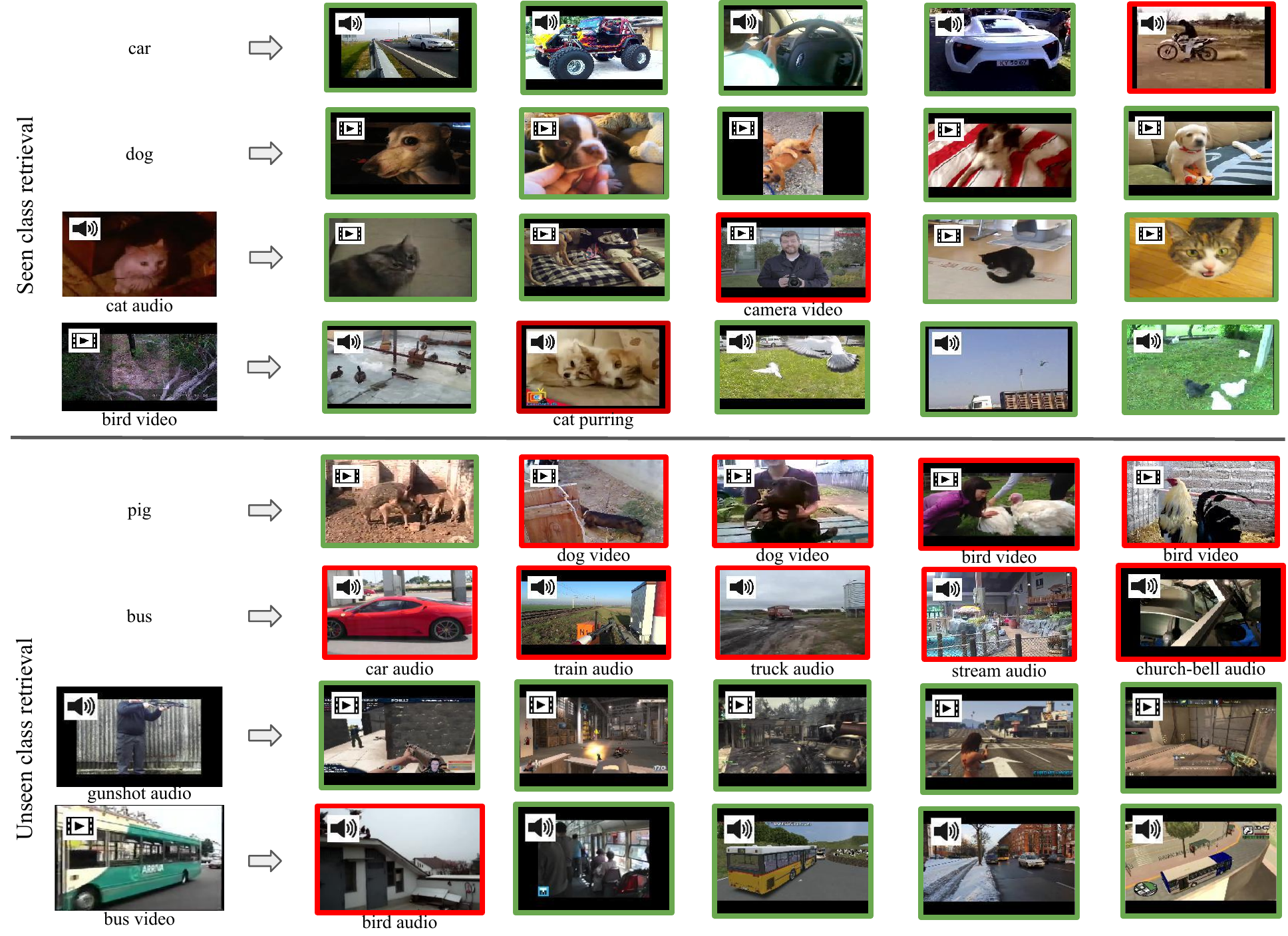}
\caption{Qualitative crossmodal retrieval results with the proposed method. Each block of two rows from top to bottom corresponds to text to audio, text to video, audio to video and video to video respectively. The small icons on the left top of each image indicates the modality considered for that specific video. Please see detailed results video in the supplementary material.}
\label{fig:qualres}
\vspace{-0.4em}
\end{figure*}
 
 \subsection{Comparison with state of the art methods}
 \label{sec:sota}
 We address both possible issues , i.e.\ (i) our implementation is competitive \wrt other methods on standard datasets, and (ii) how do other methods compare on the proposed audio-visual ZSL dataset, by providing additional results. Tab.~\ref{tab:zsl_otherdata} gives the performance of our implementation on other datasets (existing method performances are taken from Xian et al.~\cite{xian2018zero}). 
 
 We observe that our method is competitive to other methods on an average. Tab.~\ref{tab:zsl_owndata} gives the classification performance of other methods using our features on the proposed dataset. We see that our method performs better than many existing methods (\eg ALE $33.0$ vs.\ CJME $34.4$). Hence we conclude that our implementation and method, both, perform comparable to existing appearance based ZSL methods. Tab.~\ref{tab:zsl_owndata} also shows that adding audio improves the video only ZSL from $33.3$ to $34.4$ HM. 
 
 In these comparisons, we have not included some of the recent generative approaches \cite{verma2018generalized, xian2018feature} which handles the task by conditional generation of examples form the unseen classes. Although these approaches increase the performance but they come with the drawback of soft-max classification, which requires the classifier to be trained form scratch once again if a new class is added to the existing setup at test time. This also requires saving all the training data for generative approaches while in the projection based methods, this is not required.

 \subsection{Qualitative evaluation}
 Fig.~\ref{fig:qualres} shows qualitative crossmodal retrieval results for all three pairs of modalities, \ie text to audio/video, audio to video and video to audio. We see that method makes acceptable mistakes, \eg for the car text query one of the audio retrieval contains motorbike due to the similar sound, for the bird video query the wrong retrieval is a cat purring sound which is similar to a pigeon sound. In the unseen class case, the bus text query return car, train and truck audio as the top false positives. Easier and distinct cases such as gunshot audio query gives very good video retrievals. We encourage the readers to look at the result videos available at \url{https://www.cse.iitk.ac.in/users/kranti/avzsl.html} for a better understanding of qualitative results. 
\section{Conclusion}
We presented a novel method, which we call Coordinated Crossmodal Joint Embeddings (CJME), for the task of audio visual zeroshot classification and retrieval of videos. The method learns to embeds audio, video and text into a common embedding space and then performs nearest neighbor retrieval in that space for classification and retrieval. The loss function we propose has three components, two bimodal text-audio and text-video triplet losses, and an audio-video crossmodal similarity based loss. Motivated by the fact that the two modalities might carry different amount of information for different examples, we also proposed a modality attention learning framework. The attention part learns to predict the dominant modality for the task, \ie if the object is occluded but the audio is clear, and base the prediction on that modality only. We reported extensive experiments to validate the method and showed advantages of the method over baselines, as well as demonstrated crossmodal retrieval which is not possible with the baseline methods. We also constructed a dataset appropriate for the task which is a subset of a large scale unconstrained dataset for audio event detection in video. We plan to  release the dataset details upon acceptance.

{\small
\bibliographystyle{unsrt}
\bibliography{egbib}
}

\clearpage

\appendix

\section{Difference with previously claimed zeroshot video retrieval approach}
There have been some earlier works \cite{dalton2013zero, gan2016concepts} that claimed to be zeroshot video retrieval but their definition of zeroshot is different from the contemporary meaning   \cite{xian2018zero}.
In these earlier works of zeroshot, there is no separation of seen and unseen class queries, which clearly violates the contemporary meaning of zeroshot. 

The authors claim these works to be zeroshot as there is no pairing between the query and the example retrieval lists in the training set. In order to learn from the unpaired dataset, 'concept' detectors (e.g. airplanes, bicycles, church, computers) are used separately for both the modalities to extract the concepts. The extracted concepts are finally aligned for cross-modal retrieval. However, these concept detectors are pre-trained on external annotated datasets (e.g. ImageNet,  UCF101) which means that their models have already seen the concepts prior to testing.

This again violates the contemporary setting of zero-shot where no pre-training is allowed for the unseen data.

Our definition of zero-shot is aligned with the more recent and strict definition \cite{xian2018zero}, where the unseen query concept class examples are never seen during training (i.e.\ even pre-trained detectors are not allowed). Hence, our approach of zero-shot is different from the previously claimed approach and can not be compared directly with them. 

\section{Dataset}
\label{sec:data_stats}

\begin{figure*}

\centering
	\begin{tabular}{cc}
	    \includegraphics[width=\columnwidth]{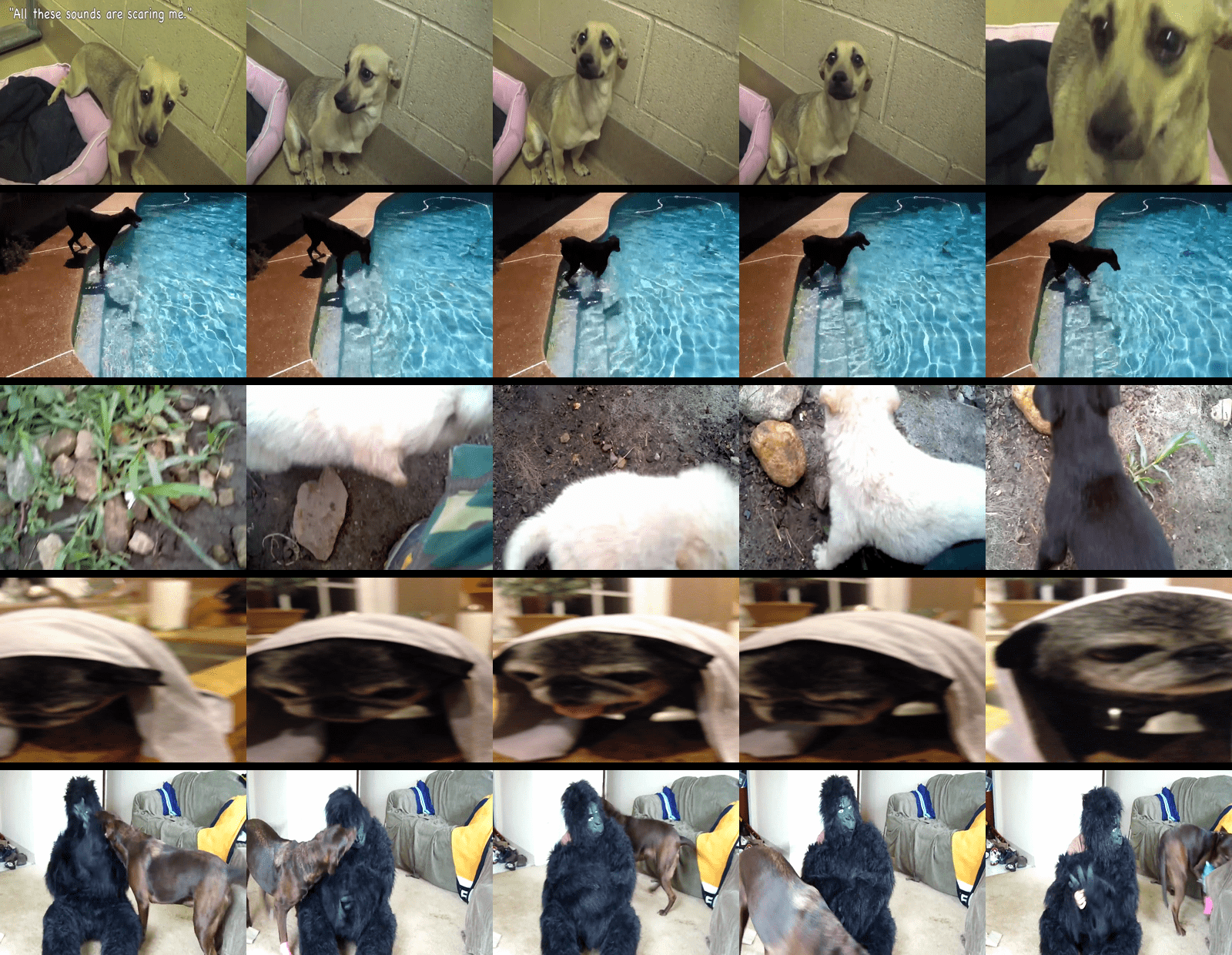}&
	    \hspace{-10pt}
		\includegraphics[width=\columnwidth]{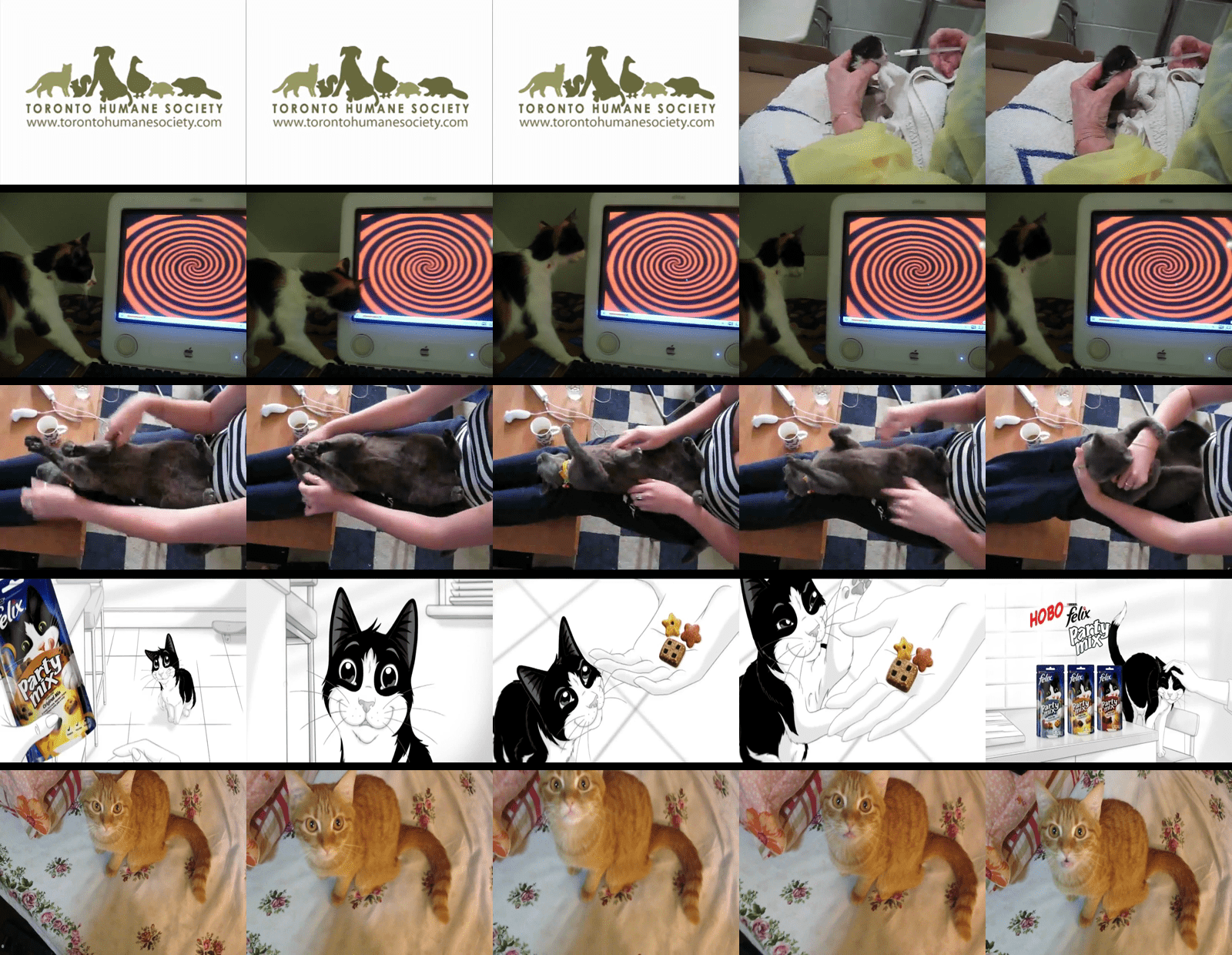}\\
		
		(Dog) & (Cat)\\

		\includegraphics[width=\columnwidth]{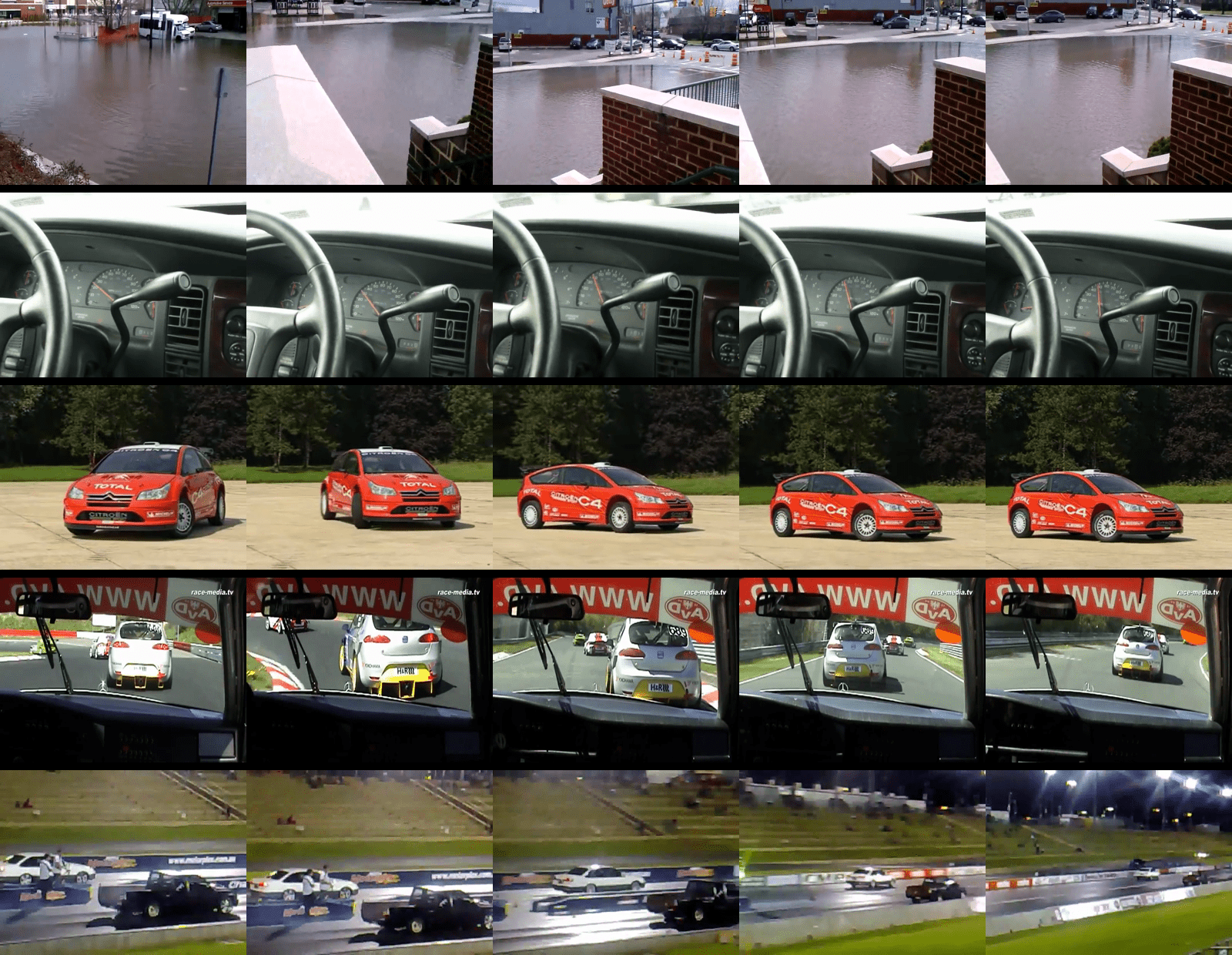}&
	    \hspace{-10pt}
		\includegraphics[width=\columnwidth]{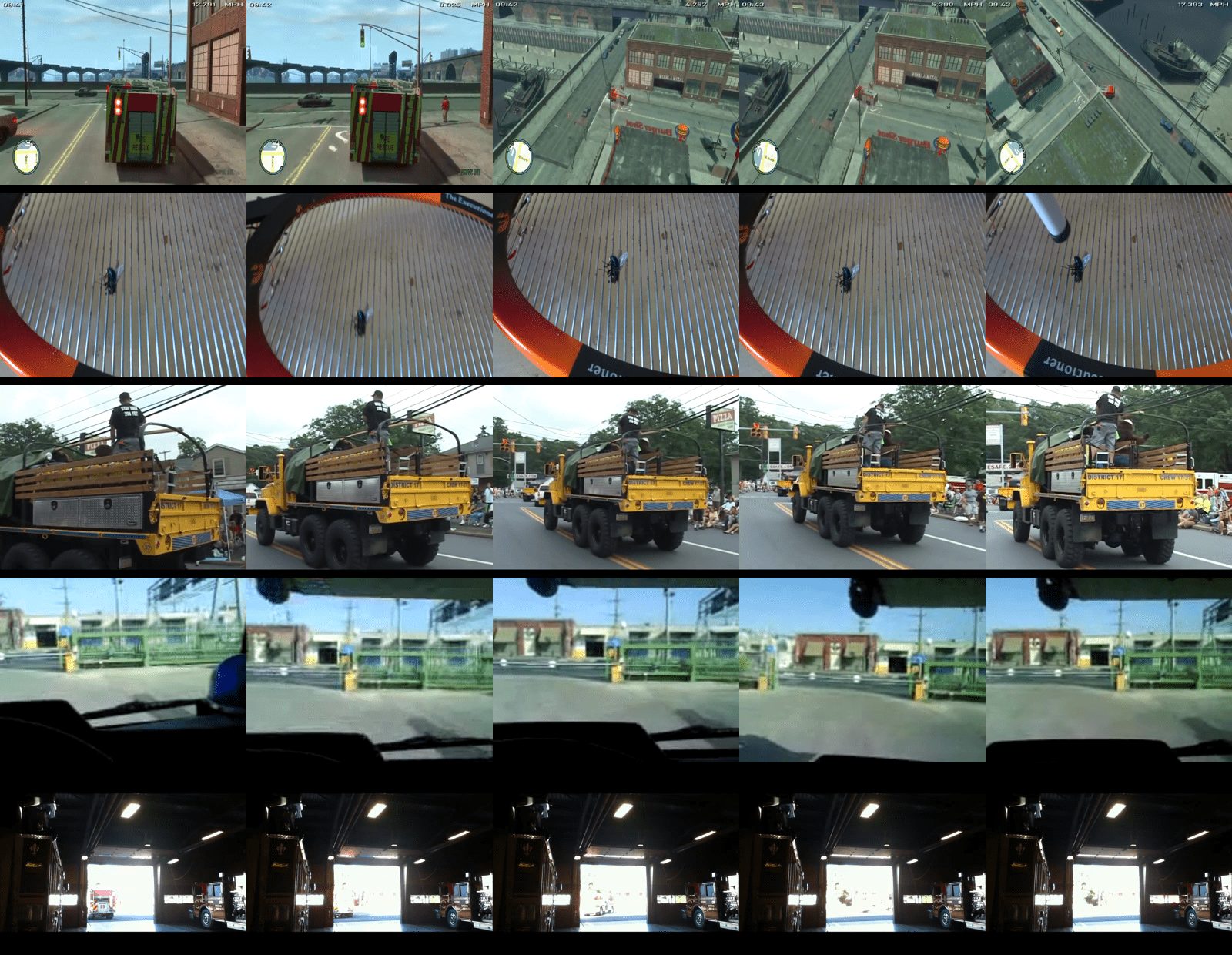}\\
		(Car) & (Ambulance)\\
		\includegraphics[width=\columnwidth]{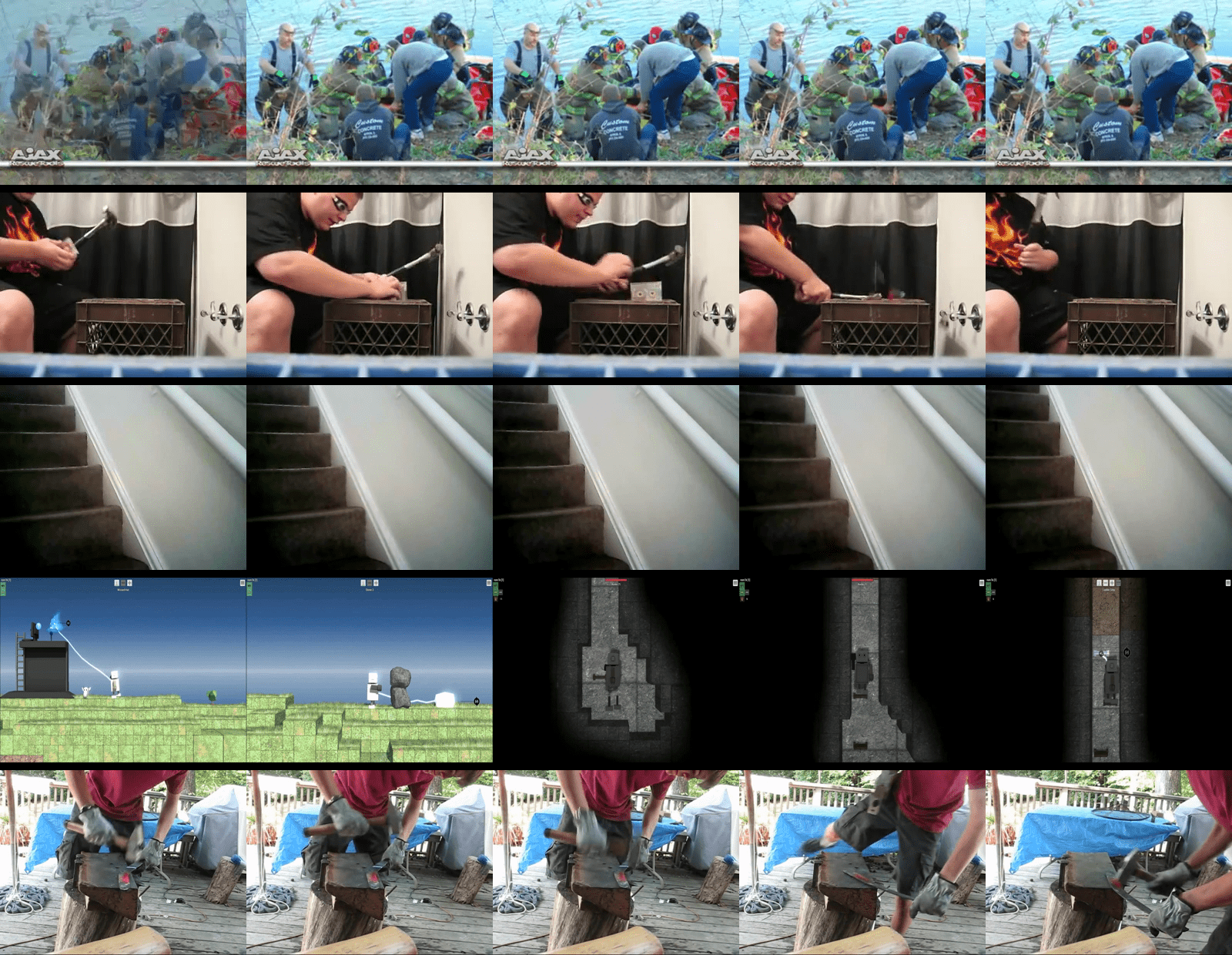}&
	    \hspace{-10pt}
		\includegraphics[width=\columnwidth]{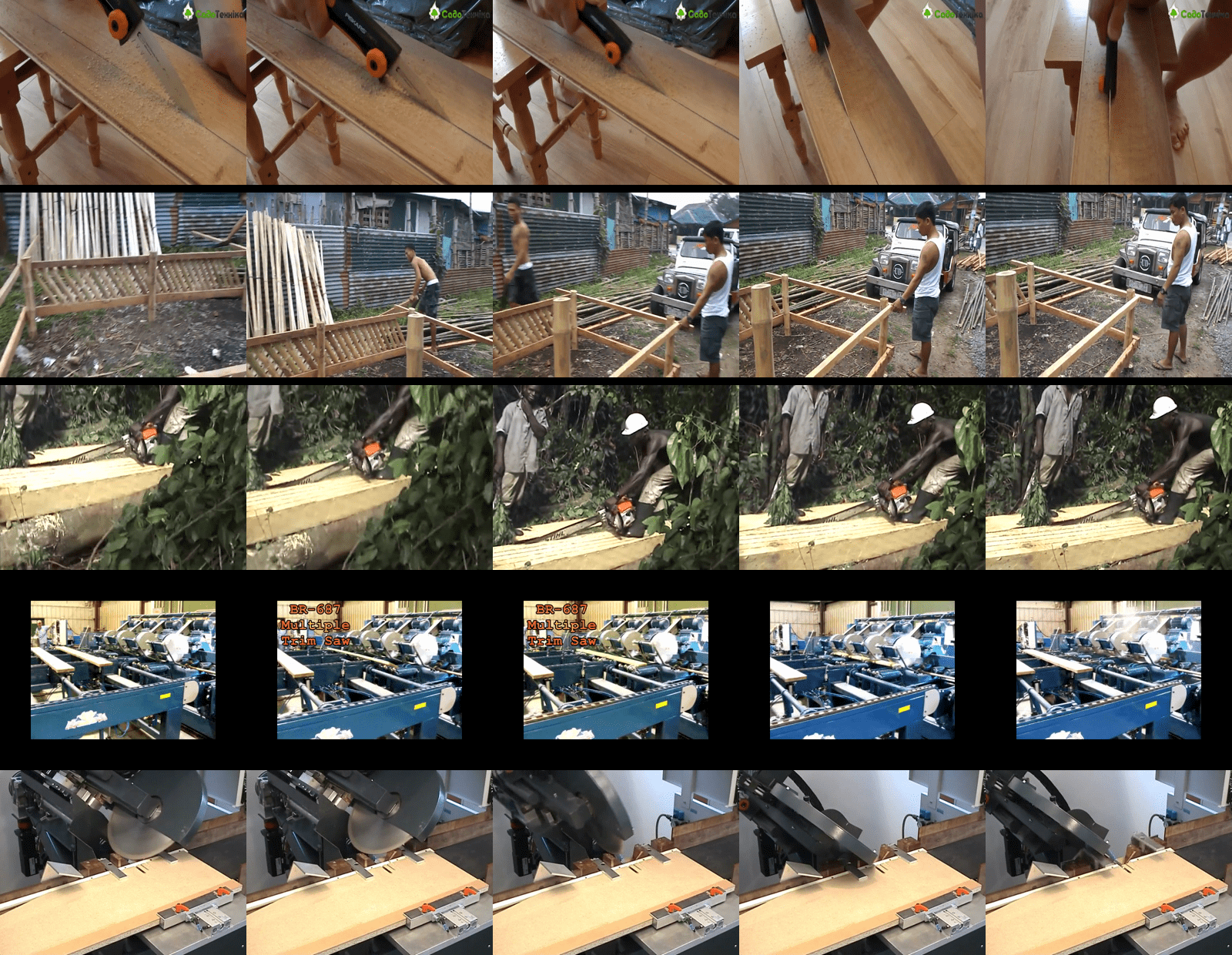}\\
		(Hammer) & (Sawing)\\

	\end{tabular}
\caption{Example videos from \textbf{seen classes} of the dataset. The classes are mentioned below each of the figure. Each row in the figure corresponds to an example video, where the frames are extracted at equal intervals from the entire video. }
\label{fig:datset_examples_seen}
\end{figure*}

\begin{figure*}

\centering
	\begin{tabular}{cc}
	    \includegraphics[width=\columnwidth]{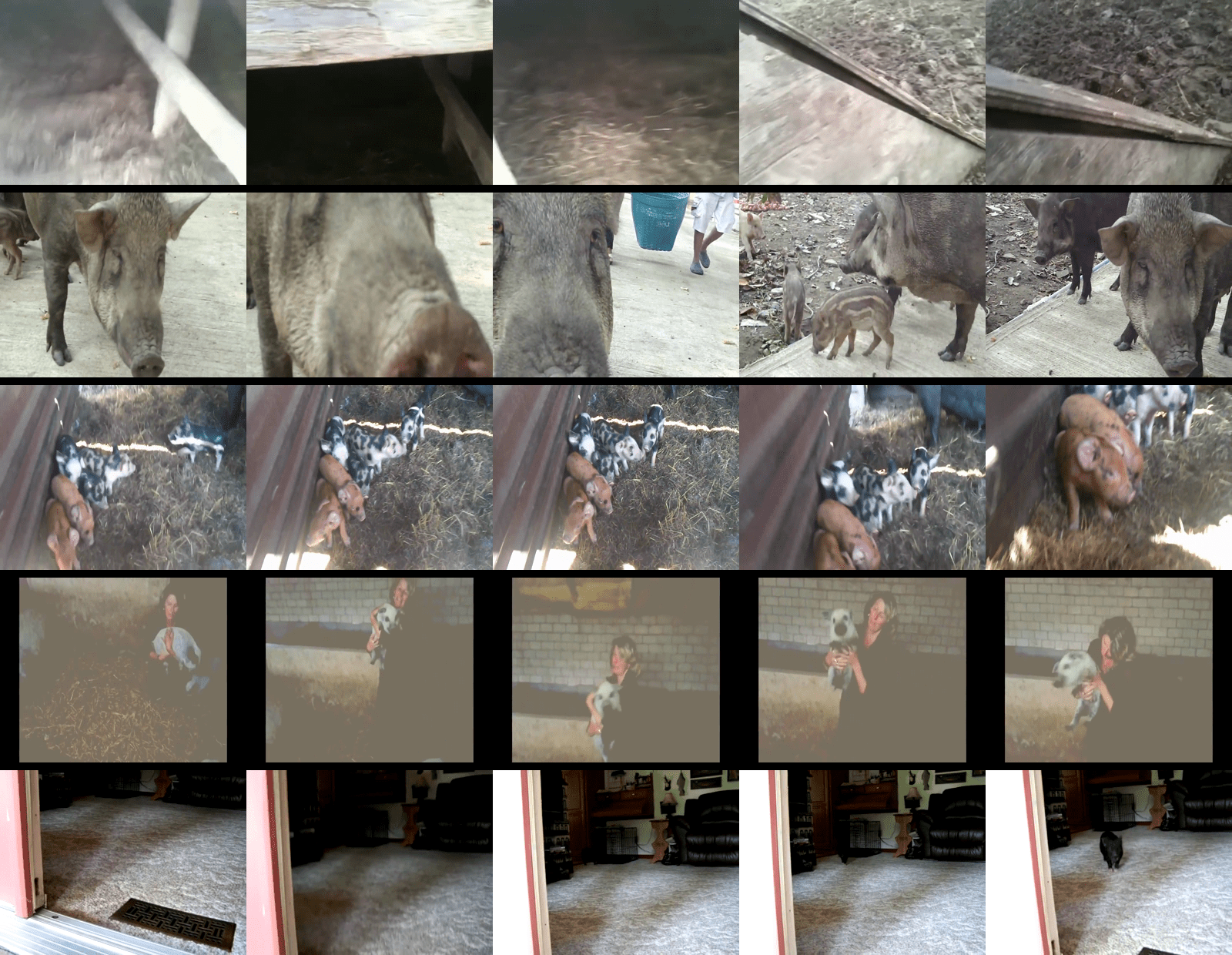}&
	    \hspace{-10pt}
		\includegraphics[width=\columnwidth]{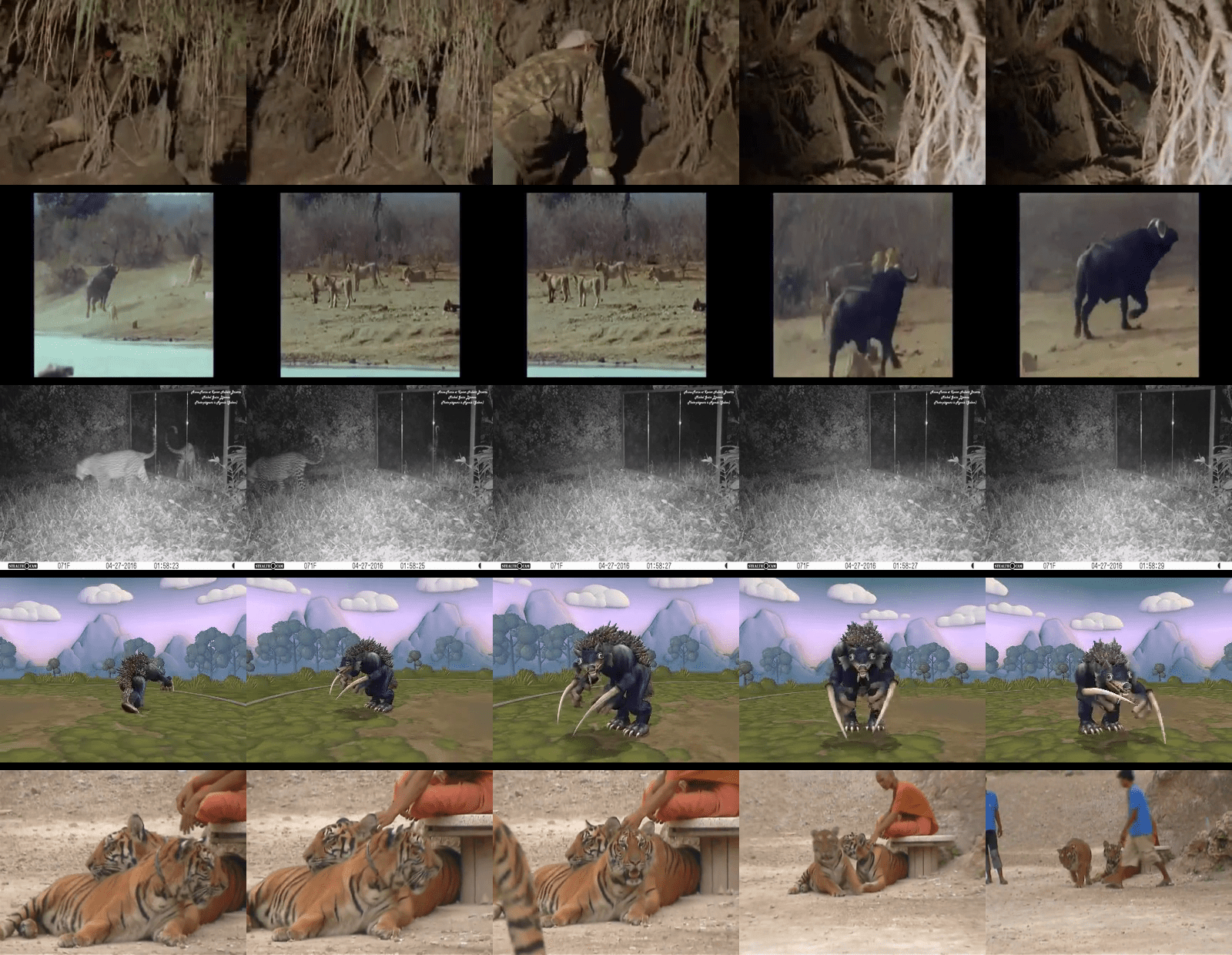}\\
		
		(Pig) & (Panther)\\

		\includegraphics[width=\columnwidth]{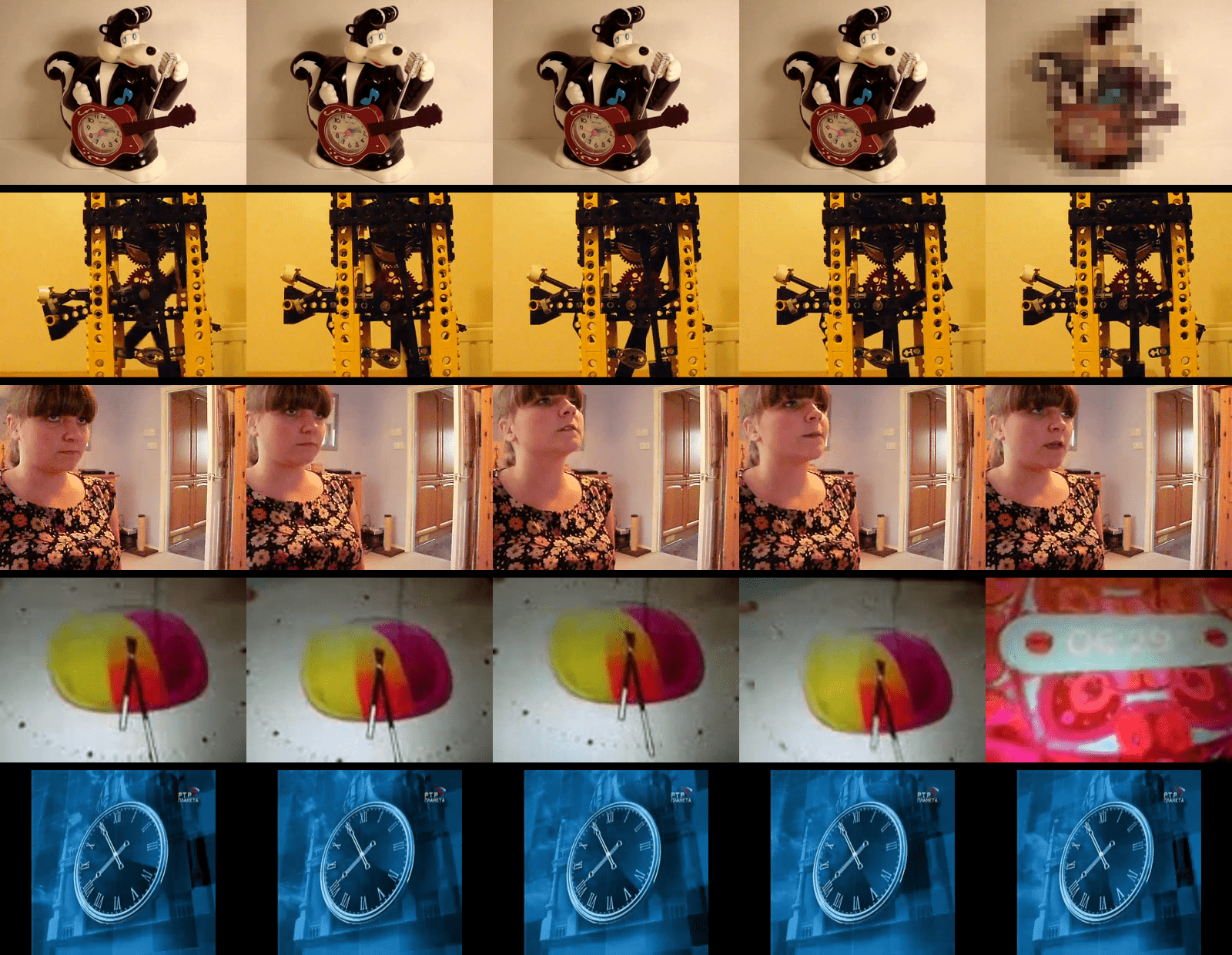}&
	    \hspace{-10pt}
		\includegraphics[width=\columnwidth]{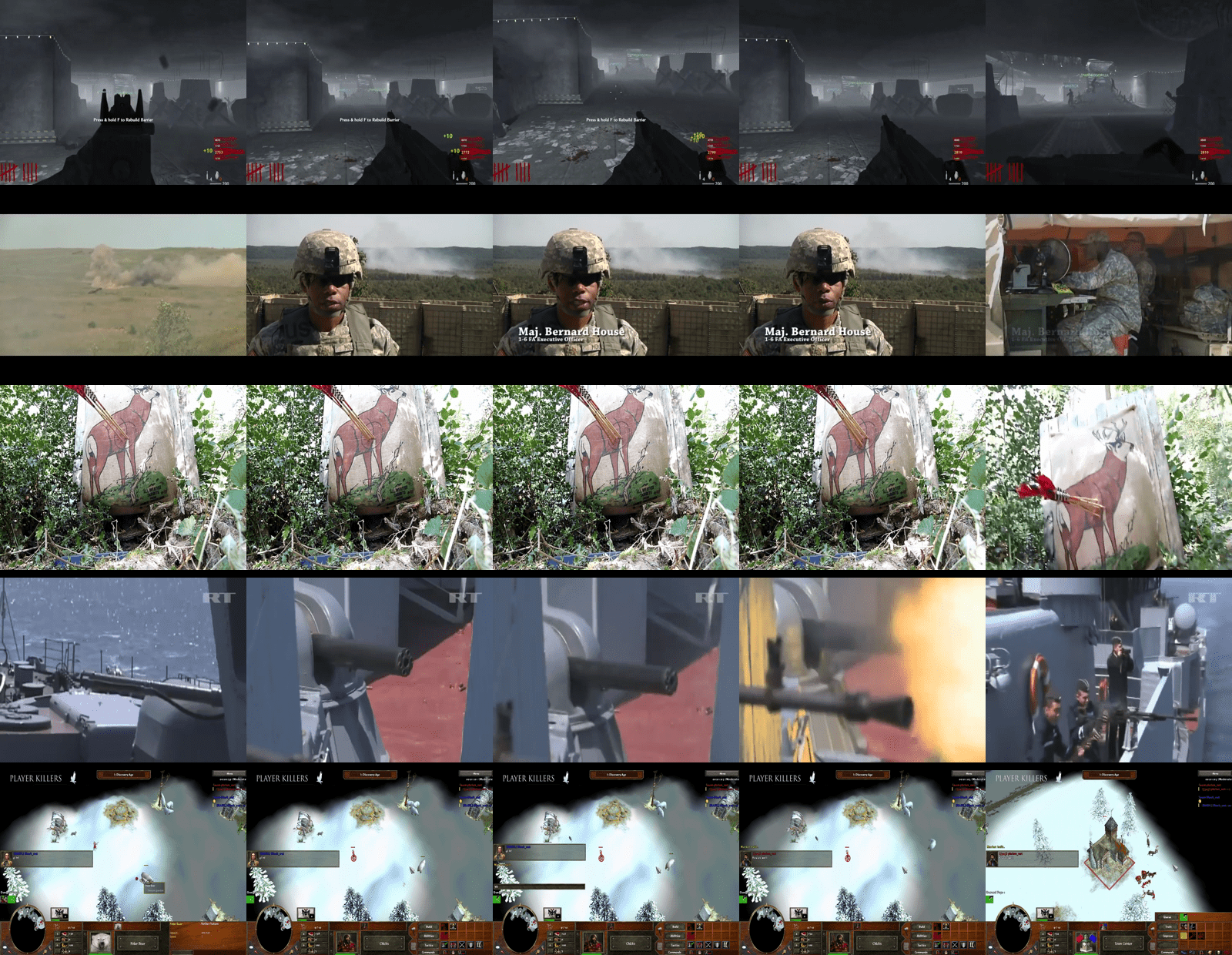}\\
		(Clock) & (Gunshot)\\
	\end{tabular}
\caption{Example videos from \textbf{unseen classes} of the dataset. The classes are mentioned below each of the figure. Each row in the figure corresponds to an example video, where the frames are extracted at equal intervals from the entire video. }
\label{fig:datset_examples_unseen}
\end{figure*}

In this section we give more details about the dataset, \texttt{AudioSetZSL} as mentioned in Section.~4 of the paper. The statistics for different splits of the dataset is given in Table.~\ref{table:dataset_stats}. The number of examples in the \textit{seen} and \textit{unseen} classes is given in Table.~\ref{table:dataset_zeroshot} and the number of examples in each class of the dataset is shown in Table.~\ref{table:dataset_classwise}.  We have provided some examples videos from the seen and unseen classes in Fig.~\ref{fig:datset_examples_seen} and Fig.~\ref{fig:datset_examples_unseen} respectively. As the dataset was collected for the audio task, it can be clearly seen that some of the frame doesn't contain the video as suggested in the paper. This can be seen for the $2^{nd}$ examples in the class \texttt{ambulance}.

\begin{table}[!ht]
\centering
\begin{tabular}{c|c|c|c|c}
\hline
split & min & max & mean & std. dev. \\
\hline  \hline
\texttt{train} & 176 & 22989 & 2844.39 & 4615.77\\
\texttt{val} & 58 & 7663 & 947.61 & 1538.65\\
\texttt{test} & 58 & 7663 & 947.88 & 1538.68 \\
\hline
\end{tabular}
\caption{Statistics on the number of examples per class for the \texttt{AudioSetZSL} dataset.}
\label{table:dataset_stats}
\end{table}

\begin{table}[!ht]
\centering
\begin{tabular}{c|c|c|c}
\hline
class & train & val & test \\
\hline  \hline
seen classes & 79795 & 26587 & 26593\\
unseen classes & 14070 & 4684 & 4687\\
\hline
\textbf{Total} & 93865 & 31271 & 31280 \\
\hline
\end{tabular}
\caption{No. of examples in seen and unseen classes of the dataset}
\label{table:dataset_zeroshot}
\end{table}

\begin{table}[t]
\centering
\begin{tabular}{c|c|c|c}
\hline
class & train & val & test \\
\hline  \hline
\texttt{dog} & 7588 & 2529 & 2529\\
\texttt{cat} & 2133 & 710 & 711\\
\texttt{horse} & 1862 & 620 & 620\\
\texttt{cattle} & 437 & 145 & 145\\
\texttt{\textbf{{pig}}} & 467 & 155 & 156\\
\texttt{goat} & 1096& 365 & 365\\
\texttt{\textbf{{panther}}} & 379 & 126 & 126\\
\texttt{bird} & 15092& 5030 & 5031\\
\texttt{thunderstorm} & 192& 64 & 64\\
\texttt{rain} & 1317& 439 & 439\\
\texttt{\textbf{{stream}}} & 1583 & 527 & 527\\
\texttt{ocean} & 1871 & 623 & 623\\
\texttt{car} & 22989& 7663 & 7663\\
\texttt{truck} & 5514& 1837 & 1838\\
\texttt{\textbf{{bus}}} & 2888& 962 & 962\\
\texttt{ambulance} & 2540& 846 & 846\\
\texttt{motorcycle} & 4000& 1333& 1333\\
\texttt{train} & 5140& 17113 & 1713\\
\texttt{\textbf{{aircraft}}} & 3104& 1034 & 1034\\
\texttt{bicycle} & 1622& 540 & 541\\
\texttt{skateboard} & 1703& 567 & 568\\
\texttt{\textbf{{clock}}} & 389& 129 & 130\\
\texttt{sewing} & 1050& 350 & 350\\
\texttt{\textbf{{fan}}} & 434& 144 & 144\\
\texttt{cashbox} & 176& 58 & 58\\
\texttt{\textbf{{printer}}} & 1915& 638 & 638\\
\texttt{camera} & 204& 68 & 68\\
\texttt{\textbf{{church-bell}}} & 662& 220 & 220\\
\texttt{hammer} & 260& 86 & 86\\
\texttt{sawing} & 431& 143 & 143\\
\texttt{\textbf{{gunshot}}} & 2249& 749 & 750\\
\texttt{fireworks} & 1699& 566 & 566\\
\texttt{boom} & 879& 292 & 293\\
\hline
\textbf{Total} & 93865 & 31271 & 31280 \\
\hline
\end{tabular}

\caption{Number of examples per class available in the \texttt{AudioSetZSL} dataset. The zeroshot classes are marked with boldface.}
\label{table:dataset_classwise}
\vspace{-1em}
\end{table}

\end{document}